\documentclass[]{llncs}

\usepackage{amssymb}  
\usepackage{epsf}     
\usepackage{multicol} 
\usepackage{makeidx}  
\usepackage{paper} 

\title{Boosting Applied to Word Sense Disambiguation}

\titlerunning{Boosting Applied to WSD}

\author{Gerard Escudero \and Llu\'{\i}s M\`arquez  
\and German Rigau\thanks{This 
research has been partially funded by the Spanish Research Department
(CICYT's BASURDE project TIC98--0423--C06) and by the Catalan
Research Department (CIRIT's consolidated research group 1999SGR-150, 
CREL's Catalan WordNet project and CIRIT's grant 1999FI 00773).}}
\authorrunning{Gerard Escudero et al.}
\institute{ 
 TALP Research Center\\
 LSI Department. Universitat Polit\`ecnica de Catalunya (UPC)\\
 Jordi Girona Salgado 1--3. 
 E-08034, Barcelona. Catalonia\\
 \{{\tt escudero,lluism,g.rigau\}@lsi.upc.es}}

\begin{document}

\mainmatter

\maketitle

\begin{abstract}
\vspace*{-2mm}
In this paper Schapire and Singer's \aABMH\ boosting algorithm is
applied to the Word Sense Disambiguation (\aWSD) problem. Initial experiments 
on a set of 15 selected polysemous words show that the boosting approach surpasses 
Naive Bayes and Exemplar--based approaches, which represent state--of--the--art 
accuracy on supervised \aWSD. In order to make boosting practical for
a real learning domain of thousands of words, several ways of
accelerating the algorithm by reducing the feature space are studied. 
The best variant, which we call \aLB, is tested on the largest
sense--tagged corpus available containing 192,800 examples of the 191 most 
frequent and ambiguous English words. Again, boosting compares
favourably to the other benchmark algorithms.

\end{abstract}

\section{Introduction}
\label{s-introduction}

{\bf Word Sense Disambiguation} (\aWSD) is the problem of assigning 
the appropriate meaning (sense) to a given word in a text or 
discourse. This meaning is distinguishable from other senses 
potentially attributable to that word.
Resolving the ambiguity of words is a central problem for language 
understanding applications and their associated tasks~\cite{ide98}, 
including, for instance, machine translation, information retrieval 
and hypertext navigation, parsing, 
spelling correction, reference resolution, automatic text summarization,
etc.
 
\aWSD\ is one of the most important open problems in the Natural Language
Processing (\aNLP) field. Despite the wide range of approaches
investigated and the large effort devoted to tackling this problem, 
it is a fact that to date no large--scale, broad coverage and highly 
accurate word sense disambiguation system has been built.

The most successful current line of research is the corpus--based
approach in which statistical or Machine Learning (\aML) algorithms have 
been applied to learn statistical models or classifiers from corpora
in order to perform \aWSD. 
Generally, supervised approaches (those that learn from a previously 
semantically annotated corpus) have obtained better results than 
unsupervised methods on small sets of selected highly ambiguous
words, or artificial pseudo--words. Many standard \aML\ algorithms 
for supervised learning have been applied, such as: 
Naive Bayes \cite{ng97a,pedersen98}, \cite{ng97a,fujii98}, 
Exemplar--based learning 
Decision Lists~\cite{yarowsky94},
Neural Networks~\cite{towell98}, etc.
Further, Mooney~\cite{mooney96} has also compared all previously cited
methods on a very restricted domain and including Decision Trees and 
Rule Induction algorithms.
Unfortunately, there have been very few direct comparisons of
alternative methods on identical test data. However, it is commonly
accepted that Naive Bayes, Neural Networks and Exemplar--based
learning represent state--of--the--art accuracy on supervised \aWSD.

Supervised methods suffer from the lack of widely available semantically 
tagged corpora, from which to construct really broad coverage systems.
This is known as the ``knowledge acquisition bottleneck''.
Ng \cite{ng97b} estimates that the manual annotation effort necessary to 
build a broad coverage semantically annotated corpus would be about 16 
man-years. This extremely high overhead for supervision and,
additionally, the also serious overhead for learning/testing many
of the commonly used algorithms when scaling to real size \aWSD\ problems,
explain why supervised methods have been seriously questioned.
 
Due to this fact, recent works have focused on 
reducing the acquisition cost as well as the need 
for supervision in corpus--based methods for \aWSD.
Consequently, the following three lines of research can be found: 
1) The design of
efficient example sampling methods~\cite{engelson96,fujii98}; 
2) The use of lexical resources, such as WordNet~\cite{miller90}, 
and WWW search engines to automatically obtain from Internet 
arbitrarily large samples of word senses~\cite{leacock98,mihalcea99}; 
3) The use of unsupervised EM--like algorithms for estimating 
the statistical model parameters~\cite{pedersen98}.
It is also our belief that this body of work, and in particular the
second line, provides enough evidence towards the ``opening'' of the
acquisition bottleneck in the near future.
For that reason, it is worth further investigating the 
application of new supervised \aML\ methods to better resolve the 
\aWSD\ problem.\bigskip

\noindent{\bf Boosting Algorithms}. The main idea of boosting
algorithms is to combine many simple and
moderately accurate hypotheses (called weak classifiers) into a
single, highly accurate classifier for the task at hand. The weak
classifiers are trained sequentially and, conceptually, each of them
is trained on the examples which were most difficult to classify by
the preceding weak classifiers.

The \aABMH\ algorithm applied in this paper~\cite{schapire98a} is a
generalization of Freund and Schapire's AdaBoost
algorithm~\cite{freund97}, which has been (theoretically and
experimentally) studied extensively and which has been shown to
perform well on standard machine--learning tasks using also standard
machine--learning algorithms as weak
learners~\cite{quinlan96,freund96,dietterich98a,bauer99}.

Regarding Natural Language (\aNL) problems, \aABMH\ has been successfully
applied to Part--of--Speech (\aPOS)\ tagging~\cite{abney99}, 
Prepositional--Phrase--attachment 
disambiguation~\cite{abney99}, and, Text 
Categorization~\cite{schapire98b} with especially good results. 

The Text Categorization domain shares several properties with the
usual settings of \aWSD, such as: very high dimensionality (typical
features consist in testing the presence/absence of concrete words),
presence of many irrelevant and highly dependent features, and the fact
that both, the learned concepts and the examples, reside very sparsely
in the feature space. Therefore, the application of \aABMH\ to 
\aWSD\ seems to be a promising choice. It has to be noted that, 
apart from the excellent results obtained on \aNL\ problems, \aABMH\
has the advantages of being theoretically well founded and  
easy to implement.\medskip

The paper is organized as follows: Section~\ref{s-algorithm} is devoted to explain
in detail the \aABMH\ algorithm. Section~\ref{s-domain} describes the domain of 
application and the initial experiments performed on a reduced set of words. 
In Section~\ref{s-practical} several alternatives are explored for 
accelerating the learning process by reducing the feature space. 
The best alternative is fully tested in Section~\ref{s-experiments}. 
Finally, Section~\ref{s-conclusions} concludes and outlines some 
directions for future work.

\section{The Boosting Algorithm \aABMH}
\label{s-algorithm}
This section describes the Schapire and Singer's \aABMH\ algorithm for
multiclass multi--label classification, using exactly the same
notation given by the authors in~\cite{schapire98a,schapire98b}.

As already said, the purpose of boosting is to
find a highly accurate classification rule by combining many {\sf weak
  hypotheses} (or weak rules), each of which may be only moderately 
accurate. It is assumed that there exists a separate procedure 
called the \aWL\ for acquiring the weak hypotheses. The boosting 
algorithm finds a set of weak hypotheses by calling the weak learner 
repeatedly in a series of $T$ rounds. These weak hypotheses are then 
combined into a single rule called the {\sf combined hypothesis}. 

Let $S=\{(x_1,Y_1),\dots,(x_m,Y_m)\}$ be the set of $m$ training
examples, where each instance $x_i$ belongs to an instance space $\X$ and
each $Y_i$ is a subset of a finite set of labels or classes $\Y$. 
The size of $\Y$ is denoted by $k=|\Y|$. 

The pseudo--code of \aABMH\ is presented in 
figure~\ref{f-abmh}.
\aABMH\ maintains an $m\!\times\!k$ matrix of weights as a distribution
$D$ over examples and labels. The goal of the \aWL\ algorithm is
to find a weak hypothesis with moderately low error with respect to 
these weights. Initially, the distribution $D_1$ is uniform, but the
boosting algorithm updates the weights on each round to force the 
weak learner to concentrate on the pairs (examples,label) which are 
hardest to predict.
\begin{figure}[htb]
\noindent\rule[0pt]{12cm}{0.4pt}\vspace*{3mm}\\
{\small
\> {\bf procedure} \aABMH\ ({\bf in}:
$S=\{(x_i,Y_i)\}_{i=1}^m$)\smallskip\\
\> {\tt \#\#\#} $S$ is the set of training examples\\
\> {\tt \#\#\#} Initialize distribution $D_1$ (for all $i$, $1\leq
i\leq m$,\, and all $l$, $1\leq l\leq k$)\smallskip\\
\>\> $D_1(i,l)=1/(mk)$\smallskip\\
\> {\bf for } $t$:=1 {\bf to} $T$ {\bf do}\smallskip\\
\>\> {\tt \#\#\#} Get the weak hypothesis $h_t: \X\times\Y\rightarrow\mathbb{R}$\smallskip\\
\>\> $h_t$ = \aWL\,($X,D_t$);\smallskip\\
\>\> {\tt \#\#\#} Update distribution $D_t$ (for all $i$, $1\leq
i\leq m$,\, and all $l$, $1\leq l\leq k$)\smallskip\\
\>\>\> $D_{t+1}(i,l)\ig {\displaystyle\frac{D_t(i,l){\rm exp}(-Y_i[l]h_t(x_i,l))}{Z_t}}$\smallskip\\
\>\> {\tt \#\#\#} $Z_t$ is a normalization factor (chosen so that
$D_{t+1}$ will be a distribution)\smallskip\\
\> {\bf end-for}\vspace*{-2mm}\\
\> {\bf return} the combined hypothesis:
   $f(x,l)={\displaystyle\sum_{t=1}^{T} h_t(x,l)}$\vspace*{-2mm}\\
\> {\bf end} \aABMH\smallskip\\
}
\noindent\rule[0pt]{12cm}{0.4pt}\vspace*{0mm}
\caption{The \aABMH\ algorithm}
\vspace*{-2mm}
\label{f-abmh}
\end{figure}  

More precisely, let $D_t$ be the distribution at round $t$, and
$h_t\,:\,\X\times\Y\rightarrow\mathbb{R}$ the weak rule acquired
according to $D_t$. The sign of $h_t(x,l)$ is interpreted as a
prediction of whether label $l$ should be assigned to example $x$ or
not. The magnitude of the prediction
$|h_t(x,l)|$ is interpreted as a measure of confidence in the 
prediction. In order to understand correctly the updating formula 
this last piece of notation should be defined. Thus, given 
$Y\subseteq\Y$ and $l\pert\Y$, 
let $Y[l]$ be +1 if $l\pert Y$ and -1 otherwise.

Now, it becomes clear that the updating function increases 
(or decreases) the weights $D_t(i,l)$ for which $h_t$ makes a good 
(or bad) prediction, and that this variation is proportional to 
$|h_t(x,l)|$.  

Note that \aWSD\ is not a multi--label classification problem since
a unique sense is expected for each word in context. In our
implementation, the algorithm runs exactly in the same way as
explained above, except that sets $Y_i$ are reduced to a unique label,
and that the combined hypothesis is forced to output a unique label, which
is the one that maximizes $f(x,l)$.

Up to now, it only remains to be defined the form of the \aWL.
Schapire and Singer~\cite{schapire98a} prove that the Hamming loss 
of the \aABMH\ algorithm on the training set\footnote{i.e. the
fraction of training examples $i$ and labels $l$ for which the sign
of $f(x_i,l)$ differs from $Y_i[l]$.} is at most $\prod_{t=1}^T Z_t$,
where $Z_t$ is the normalization factor computed on round $t$. This
upper bound is used in guiding the design of the \aWL\ algorithm,
which attempts to find a weak hypothesis $h_t$ that minimizes:
$Z_t = \sum_{i=1}^{m}\sum_{l\in\Y}\, D_t(i,l){\rm exp}
(-Y_i[l]h_t(x,l))\,$.

\subsection{Weak Hypotheses for \aWSD}
As in~\cite{abney99}, very simple weak hypotheses are used to test the
value of a boolean predicate and make a prediction based on that
value. The predicates used, which are described in section~\ref{s-corpus}, 
are of the form ``$f=v$'', where $f$ is a feature and $v$ is a value 
(e.g.: ``previous\_word = {\it hospital}'').
Formally, based on a given predicate $p$, our interest lies on weak
hypotheses $h$ which make predictions of the form:\smallskip

\indent\indent $h(x,l)=\left\{\begin{array}{ll}c_{0l}&\ {\rm if\ } p
{\rm\ holds\ in\ } x\\
                                     c_{1l}&\ {\rm otherwise}
            \end{array}\right.$\medskip

\noindent where the $c_{jl}$'s are real numbers.

For a given predicate $p$, and bearing the minimization of $Z_t$ in mind,
values $c_{jl}$ should be calculated as follows.
Let $X_1$ be the subset of examples for which the predicate $p$ holds 
and let $X_0$ be the subset of examples for which the predicate $p$
does not hold. Let $\lbd\pi\rbd$, for any predicate $\pi$, be 1 
if $\pi$ holds and 0 otherwise.
Given the current distribution $D_t$, the following real numbers are 
calculated for each possible label $l$, for $j\pert\{0,1\}$, and for 
$b\pert\{+1,-1\}$:\medskip

\indent\indent $W_b^{jl}=\sum_{i=1}^mD_t(i,l)\lbd x_i\in X_j\land
  Y_i[l]=b\rbd$\medskip

That is, $W_{+1}^{jl}$ ($W_{-1}^{jl}$) is the weight (with respect to
distribution $D_t$) of the training examples in partition $X_j$ which are
(or not) labelled by $l$.
 
As it is shown in~\cite{schapire98a}, $Z_t$ is minimized for a
particular predicate by choosing:\smallskip

\indent\indent $c_{jl}=\frac{1}{2}{\rm ln}
(\frac{W_{+1}^{jl}}{W_{-1}^{jl}})$
\medskip

\noindent These settings imply that:
\medskip

\indent\indent $Z_t=2\sum_{j\in\{0,1\}}
\sum_{l\in\Y}\sqrt{W_{+1}^{jl}W_{-1}^{jl}}$
\medskip

Thus, the predicate $p$ chosen is that for which the value of 
$Z_t$ is smallest. 

Very small or zero values for the parameters $W_b^{jl}$ cause
$c_{jl}$ predictions to be large or infinite in magnitude.
In practice, such large predictions may cause numerical problems 
to the algorithm, and seem to increase the tendency to
overfit. As suggested in~\cite{schapire98b}, smoothed values 
for $c_{jl}$ have been used. 
 
\section{Applying Boosting to \aWSD}
\label{s-domain}
%
\subsection{Corpus}
\label{s-corpus}
In our experiments the boosting approach has been evaluated using the \aDSO\ corpus
containing 192,800 semantically annotated occurrences\footnote{These examples
are tagged with a set of labels which correspond, with some minor
changes, to the senses of WordNet 1.5~\cite{ng99}.} of 121 nouns and 
70 verbs. These correspond to the most frequent and ambiguous English
words. The \aDSO\ corpus
was collected by Ng and colleagues~\cite{ng96} and it is available from
the Linguistic Data Consortium (LDC)\footnote{LDC e-mail address: 
{\tt ldc@unagi.cis.upenn.edu}}.

For our first experiments, a group of 15 words (10 nouns and 5 verbs) 
which frequently appear in the related \aWSD\ literature has been 
selected. These words are described in the left hand--side 
of table~\ref{t-quinze}. Since our goal is to acquire a classifier for each
word, each row represents a classification problem. The number of classes
(senses) ranges from 4 to 30, the number of training examples from
373 to 1,500 and the number of attributes from 1,420 to 5,181.
The \aMFS\ column on the right hand--side of table~\ref{t-quinze} shows the 
percentage of the most frequent sense for each word, i.e. the accuracy that
a naive ``Most--Frequent--Sense'' classifier would obtain. 

The binary--valued attributes used for describing the examples 
correspond to the binarization of seven
features referring to a very narrow linguistic context. 
Let ``$w_{-2}\; w_{-1}\; w\; w_{+1}\; w_{+2}$'' be the 
context of 5 consecutive words around the word $w$ to be disambiguated. 
The seven features mentioned above are exactly those used in~\cite{ng97a}: 
$w_{-2}$, $w_{-1}$, $w_{+1}$, $w_{+2}$, $(w_{-2},w_{-1})$, ($w_{-1},w_{+1}$), 
and $(w_{+1},w_{+2})$, where the last three correspond to collocations of 
two consecutive words.

\subsection{Benchmark Algorithms and Experimental Methodology}
\label{ss-expmeth}
\aABMH\ has been compared to the following algorithms:\medskip\\
  \noindent{\bf Naive Bayes} (\aNB).\ \ The naive Bayesian classifier has been
  used in its most classical setting~\cite{duda73}. To avoid the effect of
  zero counts when estimating the conditional probabilities of the
  model, a very simple smoothing technique has been used, which was 
  proposed in~\cite{ng97a}.\smallskip

  \noindent{\bf Exemplar--based learning} (\aEBk).\ \ In our
  implementation, all examples are stored in memory and the 
  classification of a new example is based on a $k$--NN algorithm
  using Hamming distance to measure closeness (in doing so, all 
  examples are examined). If $k$ is greater than 1,
  the resulting sense is the weighted majority sense of the $k$ nearest 
  neighbours (each example votes its sense with a strength proportional to its
  closeness to the test example). 
  Ties are resolved in favour of the most frequent sense among all those tied.
  \smallskip

The comparison of algorithms has been performed in series 
of controlled experiments using exactly the same training and 
test sets for each method. 
The experimental methodology consisted in a 10-fold cross-validation.
All accuracy/error rate figures appearing in the paper are averaged 
over the results  of the 10 folds. The statistical tests of 
significance have been 
performed using a 10-fold cross validation paired Student's 
$t$-test with a confidence value of: $t_{9,0.975}=2.262$.

\subsection{Results}
\label{ss-results}
Figure~\ref{f-lc15} shows the error rate curve of \aABMH, averaged over 
the 15 reference words, and for an increasing number of weak rules per word. 
This plot shows that the error obtained by \aABMH\ is lower
than those obtained by \aNB\ and \aEBx\ ($k$=15 is the best choice for 
that parameter from a number of tests between $k$=1 and $k$=30) for a number 
of rules above 100. It also shows that the error rate decreases
slightly and monotonically, as it approaches the maximum number of 
rules reported\footnote{The maximum number of rounds considered is
750, merely for efficiency reasons.}.

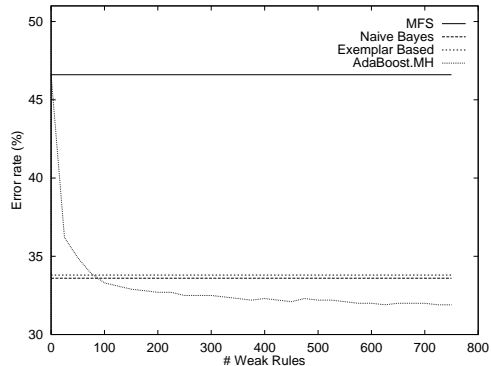
\figuraCX{lc15}{7}{Error rate of \aABMH\ related to the number of weak rules}{f-lc15}

According to the plot in figure~\ref{f-lc15}, no
overfitting is observed while increasing the number of rules per
word. Although it seems that the best strategy could be ``learn as many
rules as possible'', in~\cite{escudero00} it is shown that the number 
of rounds must be determined individually for each word since they have
different behaviours.
The adjustment of the number of rounds can be done by cross--validation
on the training set, as suggested in~\cite{abney99}. However, in our case, 
this cross--validation inside the cross--validation of the general
experiment would generate a prohibitive overhead. Instead, a very
simple stopping criterion ({\it sc}) has been used, which consists in stopping
the acquisition of weak rules whenever the error rate on the training
set falls below 5\%, with an upper bound of 750 rules. This variant, 
which is referred to as \aABsc, obtained comparable results to \aAB\ 
but generating only 370.2 weak rules per word on average, which 
represents a very moderate storage requirement for the combined classifiers.

The numerical information corresponding to this experiment is included 
in table~\ref{t-quinze}. This table shows the accuracy results, detailed for 
each word, of \aNB, \aEBi, \aEBx, \aAB, and \aABsc. The best result for each 
word is printed in boldface. 

\begin{table}[htb]
\begin{center}
\begin{tabular}{lcccccccccc} \hline
& & \multicolumn{3}{c}{{\sf Number of}} &
\multicolumn{6}{c}{{\sf Accuracy (\%)}}\\\cline{3-5}\cline{6-11} 
\ {\sf Word} &{\sf POS}\ \ & {\sf Senses} &  {\sf Examp.} & \multicolumn{1}{c|}{{\sf Attrib.}} &
\ {\sf MFS} &\ {\sf NB} &\ {\sf EB$_1$} &\ {\sf EB$_{15}$} &\ {\sf AB$_{750}$} &\ {\sf AB$_{sc}$}\\\hline
\ {\sf age}     & {\sf n} &  4 &  493 &\multicolumn{1}{c|}{1662} &\
62.1 &\ 73.8 &\ 71.4 &\ 71.0 &\ {\bf 74.7} &\ 74.0\\
\ {\sf art}     & {\sf n} &  5 &  405 &\multicolumn{1}{c|}{1557} &\
46.7 &\ 54.8 &\ 44.2 &\ 58.3 &\ 57.5 &\ {\bf 62.2}\\
\ {\sf car}     & {\sf n} &  5 & 1381 &\multicolumn{1}{c|}{4700} &\
95.1 &\ 95.4 &\ 91.3 &\ 95.8 &\ {\bf 96.8} &\ 96.5\\
\ {\sf child}   & {\sf n} &  4 & 1068 &\multicolumn{1}{c|}{3695} &\
80.9 &\ 86.8 &\ 82.3 &\ 89.5 &\ {\bf 92.8} &\ 92.2\\
\ {\sf church}  & {\sf n} &  4 &  373 &\multicolumn{1}{c|}{1420} &\
61.1 &\ 62.7 &\ 61.9 &\ 63.0 &\ {\bf 66.2} &\ 64.9\\
\ {\sf cost}    & {\sf n} &  3 & 1500 &\multicolumn{1}{c|}{4591} &\
87.3 &\ 86.7 &\ 81.1 &\ 87.7 &\ 87.1 &\ {\bf 87.8}\\
\ {\sf fall}    & {\sf v} & 19 & 1500 &\multicolumn{1}{c|}{5063} &\
70.1 &\ 76.5 &\ 73.3 &\ 79.0 &\ {\bf 81.1} &\ 80.6\\
\ {\sf head}    & {\sf n} & 14 &  870 &\multicolumn{1}{c|}{2502} &\
36.9 &\ 76.9 &\ 70.0 &\ 76.9 &\ {\bf 79.0} &\ {\bf 79.0}\\
\ {\sf interest}\ & {\sf n} &  7 & 1500 &\multicolumn{1}{c|}{4521} &\
45.1 &\ 64.5 &\ 58.3 &\ 63.3 &\ {\bf 65.4} &\ 65.1\\
\ {\sf know}    & {\sf v} &  8 & 1500 &\multicolumn{1}{c|}{3965} &\
34.9 &\ 47.3 &\ 42.2 &\ 46.7 &\ {\bf 48.7} &\ {\bf 48.7}\\
\ {\sf line}    & {\sf n} & 26 & 1342 &\multicolumn{1}{c|}{4387} &\
21.9 &\ 51.9 &\ 46.1 &\ 49.7 &\ {\bf 54.8} &\ 54.5\\
\ {\sf set}     & {\sf v} & 19 & 1311 &\multicolumn{1}{c|}{4396} &\
36.9 &\ {\bf 55.8} &\ 43.9 &\ 54.8 &\ {\bf 55.8} &\ {\bf 55.8}\\
\ {\sf speak}   & {\sf v} &  5 &  517 &\multicolumn{1}{c|}{1873} &\
69.1 &\ {\bf 74.3} &\ 64.6 &\ 73.7 &\ 72.2 &\ 73.3\\
\ {\sf take}    & {\sf v} & 30 & 1500 &\multicolumn{1}{c|}{5181} &\
35.6 &\ 44.8 &\ 39.3 &\ 46.1 &\ {\bf 46.7} &\ 46.1\\
\ {\sf work}    & {\sf n} &  7 & 1469 &\multicolumn{1}{c|}{4923} &\
31.7 &\ {\bf 51.9} &\ 42.5 &\ 47.2 &\ 50.7 &\ 50.7\\\hline
\multicolumn{2}{l}{\ Avg. nouns}& 8.6 & 1040.1
&\multicolumn{1}{c|}{\ 3978.5\ \ } &\ 57.4 &\ 71.7 &\ 65.8 &\ 71.1 &\ {\bf 73.5} &\ 73.4\\
\multicolumn{2}{l}{\ \phantom{Avg.} verbs}& 17.9 & 1265.6 
&\multicolumn{1}{c|}{\ 4431.9\ \ } &\ 46.6 &\ 57.6 &\ 51.1 &\ 58.1 &\ {\bf 59.3} &\ 59.1\\\hline
\multicolumn{2}{l}{\ \phantom{Avg.} all}& 12.1 & 1115.3
&\multicolumn{1}{c|}{\ 4150.0\ \ } &\ 53.3 &\ 66.4 &\ 60.2 &\ 66.2 &\ {\bf 68.1} &\ 68.0\\\hline
\end{tabular}
\end{center}
\caption{Set of 15 reference words and results of the main algorithms}
\vspace*{-5mm}\label{t-quinze}
\end{table}

As it can be seen, in 14 out of 15 cases, the best results correspond to the 
boosting algorithms. When comparing global results, accuracies of either
\aAB\ or \aABsc\ are significantly greater than those of any of the other 
methods. Finally, note that accuracies corresponding to \aNB\ and 
\aEBx\ are comparable (as suggested in~\cite{ng97a}), and that the use
of $k$'s greater than 1 is crucial for making Exemplar--based learning 
competitive on \aWSD.

\section{Making Boosting Practical for \aWSD} 
\label{s-practical}
Up to now, it has been seen that \aABMH\ is a simple and competitive
algorithm for the \aWSD\ task. It achieves an accuracy performance
superior to that of the Naive Bayes and Exemplar--based algorithms
tested in this paper.    
However, \aABMH\ has the drawback of its computational 
cost, which makes the algorithm not scale properly to 
real \aWSD\ domains of thousands of words.

The space and time--per--round requirements of \aABMH\ are  
$\mathcal{O}(mk)$ (recall that $m$ is the number of training examples
and $k$ the number of senses), not including the call to the weak learner.
This cost is unavoidable since \aABMH\ is inherently sequential. That
is, in order to learn the ($t$+1)-th weak rule it needs the calculation of
the $t$-th weak rule, which properly updates the matrix $D_{t}$.
Further, inside the \aWL, there is another iterative process that examines,
one by one, all attributes so as to decide which is the one that minimizes
$Z_t$. Since there are thousands of attributes, this is also a time
consuming part, which can be straightforwardly spedup either by
reducing the number of attributes or by relaxing the need to 
examine all attributes at each iteration.

\subsection{Accelerating the \aWL}
Four methods have been tested in order to reduce the cost of 
searching for weak rules. The first three, consisting in aggressively 
reducing the feature space, are frequently applied in Text Categorization.
The fourth consists in reducing the number of attributes that are 
examined at each round of the boosting algorithm.\medskip\\
  {\bf Frequency filtering} ({\sf Freq}):\ This method consists in simply
  discarding  those features corresponding to events that occur less 
  than $N$ times in the training corpus. The idea beyond that
  criterion is that frequent events are more informative than rare 
  ones.\smallskip\\
  {\bf Local frequency filtering} ({\sf LFreq}):\ This method works similarly
  to {\sf Freq} but considers the frequency of events locally, at
  the sense level. More particularly, it selects the $N$ most frequent features 
  of each sense.\smallskip\\
  {\bf RLM ranking}:\ This third method consists in making a 
  ranking of all attributes according to the \aRLM\ distance measure
  \cite{lopez91} and selecting the $N$ most relevant features. 
  This measure has been commonly used for attribute selection in 
  decision tree induction algorithms\footnote{\aRLM\ distance 
  belongs to the distance--based and information--based 
  families of attribute selection functions. It has been selected 
  because it showed better performance than seven other alternatives 
  in an experiment of decision tree induction for \aPOS\ 
  tagging~\cite{marquez99}.}.\smallskip\\
  {\bf LazyBoosting}:\ The last method does not filter out any attribute but 
  reduces the number of those that are examined at each iteration of the
  boosting algorithm. More specifically, a small proportion $p$ of 
  attributes are randomly selected and the best weak rule is selected among them.
  The idea behind this method is that if the proportion $p$ is not too small,
  probably a sufficiently good rule can be found at each iteration.
  Besides, the chance for a good rule to appear in the whole
  learning process is very high. Another important characteristic is that
  no attribute needs to be discarded and so we avoid the risk of
  eliminating relevant attributes\footnote{This 
  method will be called \aLB\ in reference to the work by Samuel 
  and colleagues~\cite{samuel98}. They applied the same technique 
  for accelerating the learning algorithm in a Dialogue Act tagging 
  system.}.\medskip

The four methods above have been compared for the set of 15 reference words.
Figure~\ref{f-ap250} contains the average error--rate curves obtained 
by the four variants at increasing levels of attribute reduction. The top
horizontal line corresponds to the \aMFS\ error rate, while the bottom
horizontal line stands for the error rate of \aABMH\ working with all
attributes. 
The results contained in figure \ref{f-ap250} are calculated running
the boosting algorithm 250 rounds for each word. 

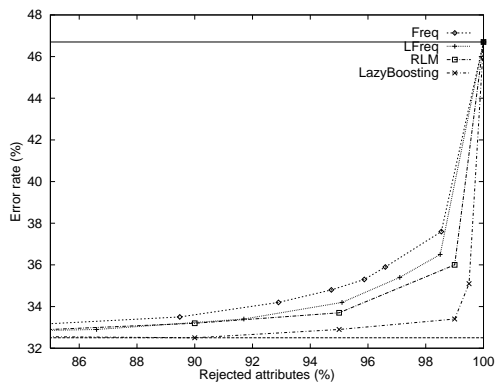
\figuraCX{ap250}{7.1}{Error rate obtained by the four methods, at 250 weak
  rules per word, with respect to the percentage of rejected
  attributes}{f-ap250}

The main conclusions that can be drawn are the following:
\begin{itemize}
\item[$\bullet$] All methods seem to work quite well since no important
  degradation is observed in performance for values lower than
  95\% in rejected attributes. This may indicate that there are 
  many irrelevant or highly dependent attributes in our domain.
\item[$\bullet$] {\sf LFreq} is slightly better than {\sf Freq}, indicating
  a preference to make frequency counts for each sense rather than
  globally.
\item[$\bullet$] The more informed \aRLM\ ranking performs better than 
  frequency--based reduction methods {\sf Freq} and {\sf LFreq}.
\item[$\bullet$] \aLB\ is better than all other methods, confirming our expectations:
  it is worth keeping all information provided by the features.
  In this case, acceptable performance is obtained even if
  only 1\% of the attributes is explored when looking for a weak 
  rule. The value of 10\%, for which \aLB\ still achieves the same 
  performance and runs about 7 times faster than \aABMH\ working with 
  all attributes, will be selected for the experiments in section~\ref{s-experiments}.
\end{itemize}

\section{Evaluating \aLB}
\label{s-experiments}
The \aLB\ algorithm has been tested on the full semantically annotated
corpus with $p=10\%$ and the same stopping criterion described in 
section~\ref{ss-results}, which will be referred to as \aABlsc.
The average number of senses is 7.2 for nouns, 12.6 for verbs,
and 9.2 overall. The average number of training examples is 933.9 for
nouns, 938.7 for verbs, and 935.6 overall. 

The \aABlsc\ algorithm learned an average of 381.1 rules per word, 
and took about 4 days of {\sc cpu} time to 
complete\footnote{The current implementation is written in PERL-5.003
  and it was run on a SUN UltraSparc2 machine with 194Mb of
  RAM.}. 
It has to be noted that this time includes the cross--validation overhead.
Eliminating it, it is estimated that 4 {\sc cpu} days would be the necessary
time for acquiring a word sense disambiguation boosting--based system
covering about 2,000 words. 

The \aABlsc\ has been compared again to the benchmark algorithms
using the 10-fold cross--validation methodology described in section 
\ref{ss-expmeth}.
The average accuracy results are reported in the left hand--side of 
table~\ref{t-cnu}. The best figures correspond to the \aLB\ algorithm
\aABlsc , and again, the differences are statistically significant using the
10-fold cross--validation paired $t$-test. 
\begin{table}[htb]
\begin{center}
\begin{tabular}{lcccc|cc} \hline
 & \multicolumn{4}{c}{{\sf  Accuracy (\%)}} &
  \multicolumn{2}{c}{{\sf Wins--Ties--Losses}}\\\cline{2-5}\cline{6-7} 
 &\ \ {\sf MFS} &\ {\sf NB} &\ {\sf EB$_{15}$} &\ {\sf AB$_{l10sc}$}\ \ &\ \
 {\sf AB$_{l10sc}$ vs. {\sf NB}} &\ \ \ {\sf AB$_{l10sc}$ vs. {\sf  EB$_{15}$}}\\\hline
\multicolumn{1}{l|}{\ Nouns (121)}  &\ 56.4 &\ 68.7 &\ 68.0 & {\bf
  70.8} &\ 99(51)\,--\,1\,--\,21(3) &\ 100(68)\,--\,5\,--\,16(1)\\
\multicolumn{1}{l|}{\ Verbs (70)}   &\ 46.7 &\ 64.8 &\ 64.9 & {\bf
  67.5} &\ 63(35)\,--\,1\,--\,6(2) &\ 64(39)\,--\,2\,--\,4(0)\\\hline
\multicolumn{1}{l|}{\ Average (191)}&\ 52.3 &\ 67.1 &\ 66.7 & {\bf
  69.5} &\ 162(86)\,--\,2\,--\,27(5) &\ 164(107)\,--\,7\,--\,20(1)\\\hline
\end{tabular}
\end{center}
\caption{Results of \aLB\ and the benchmark methods on the 191--word corpus}
\vspace*{-5mm}\label{t-cnu}
\end{table}

The right hand--side of the table shows the comparison of \aABlsc\ versus
\aNB\ and \aEBx\ algorithms, respectively.
Each cell contains the number of {\sf wins}, {\sf ties}, and {\sf
  losses} of competing algorithms. The counts of statistically
significant differences are included in brackets.
It is important to point out that \aEBx\ only beats significantly \aABlsc\
in one case while \aNB\ does so in five cases. Conversely, a significant 
superiority of \aABlsc\ over \aEBx\ and \aNB\ is observed in 107 and 86
cases, respectively.

\section{Conclusions and Future Work}
\label{s-conclusions}
In the present work, Schapire and Singer's \aABMH\ algorithm has been evaluated
on the word sense disambiguation task, which is one of the hardest open problems
in Natural Language Processing. As it has been shown, the boosting approach outperforms
Naive Bayes and Exemplar--based learning, which represent state--of--the--art accuracy
on supervised \aWSD. In addition, a faster variant has been suggested
and tested, which is called \aLB. This variant allows the scaling of the algorithm
to broad-coverage real \aWSD\ domains, and is as accurate as \aABMH.
Further details can be found in an extended version of this 
paper~\cite{escudero00}.

Future work is planned to be done in the following directions:
\begin{itemize}
\item[$\bullet$] Extensively evaluate \aABMH\ on the \aWSD\ task. 
This would include taking into account additional attributes, and 
testing the algorithms in other manually annotated corpora, and 
especially on sense--tagged corpora automatically obtained from Internet. 
\item[$\bullet$] Confirm the validity of the \aLB\ approach on other 
language learning tasks in which \aABMH\ works well, e.g.: Text Categorization.
\item[$\bullet$] It is known that mislabelled examples resulting from 
annotation errors tend to be hard examples to classify correctly, and,
therefore, tend to have large weights in the final distribution. This 
observation allows both to identify the noisy examples and use
boosting as a way to improve data quality~\cite{schapire98b,abney99}. 
It is suspected that the corpus used in the current work is very
noisy, so it could be worth using boosting to try and improve it.  
\end{itemize} 


\end{document}